\documentclass{sigchi}

\usepackage{balance}       
\usepackage{graphics}      
\usepackage[T1]{fontenc}   
\usepackage{txfonts}
\usepackage{mathptmx}
\usepackage[pdflang={en-US},pdftex]{hyperref}
\usepackage{color}
\usepackage{booktabs}
\usepackage{textcomp}

\usepackage{microtype}        
\usepackage{ccicons}          

\usepackage{todonotes}

\def\plaintitle{SimplerVoice: A Key Message \& Visual Description Generator System for Illiteracy}
\def\emptyauthor{}
\def\plainkeywords{Authors' choice; of terms; separated; by
  semicolons; include commas, within terms only; required.}

\makeatletter
\def\url@leostyle{%
  \@ifundefined{selectfont}{
    \def\UrlFont{\sf}
  }{
    \def\UrlFont{\small\bf\ttfamily}
  }}
\makeatother
\def\pprw{8.5in}
\def\pprh{11in}

\definecolor{linkColor}{RGB}{6,125,233}
\hypersetup{%
  pdftitle={\plaintitle},
  pdfauthor={\emptyauthor},
  pdfkeywords={\plainkeywords},
  pdfdisplaydoctitle=true, 
  bookmarksnumbered,
  pdfstartview={FitH},
  colorlinks,
  citecolor=black,
  filecolor=black,
  linkcolor=black,
  urlcolor=linkColor,
  breaklinks=true,
  hypertexnames=false
}

\begin{document}

\title{\plaintitle\vspace{1.2cm}}

\numberofauthors{5}
\author{%
  \alignauthor{Minh N. B. Nguyen\\
    \affaddr{Univ. of Southern California}\\
     \affaddr{Los Angeles, CA}\\
    \email{minhnngu@usc.edu}}\\
  \alignauthor{Samuel Thomas\\
    \affaddr{IBM Research AI}\\
    \affaddr{Yorktown Heights, NY}\\
    \email{sthomas@us.ibm.com}}\\
  \alignauthor{Anne E. Gattiker\\
    \affaddr{IBM Research}\\
    \affaddr{Austin, TX}\\
    \email{gattiker@us.ibm.com}}\\
  \alignauthor{Sujatha Kashyap\\
    \affaddr{IBM Research}\\
    \affaddr{Austin, TX}\\
    \email{kashyap@us.ibm.com}}\\
  \alignauthor{Kush R. Varshney\\
    \affaddr{IBM Research AI}\\
    \affaddr{Yorktown Heights, NY}\\
    \email{krvarshn@us.ibm.com}}\\
}

\maketitle

\begin{abstract}
We introduce SimplerVoice: a key message and visual description generator system to help low-literate adults navigate the information-dense world with confidence, on their own. SimplerVoice can automatically generate sensible sentences describing an unknown object, extract semantic meanings of the object usage in the form of a query string, then, represent the string as multiple types of visual guidance (pictures, pictographs, etc.). We demonstrate SimplerVoice system in a case study of generating grocery products' manuals through a mobile application. To evaluate, we conducted a user study on SimplerVoice's generated description in comparison to the information interpreted by users from other methods: the original product package and search engines' top result, in which SimplerVoice achieved the highest performance score: 4.82 on 5-point mean opinion score scale. Our result shows that SimplerVoice is able to provide low-literate end-users with simple yet informative components to help them understand how to use the grocery products, and that the system may potentially provide benefits in other real-world use cases.
\end{abstract}



\section{Introduction}
\label{sec:intro}
Illiteracy has been one of the most serious pervasive problems all over the world. According to the U.\ S.\ Department of Education, the National Center for Education Statistics, approximately 32 million adults in the United States are not able to read, which is about 14\% of the entire adult population \cite{naal}. Additionally, 44\% of the 2.4 million students in the U.\ S.\ federally funded adult education programs  are English as a second language (ESL) students, and about 185,000 of them are at the lowest ESL level, beginning literacy \cite{condelli2010impact}. While low-literate adults lack the ability to read and to understand text, particularly, the low-literate ESL adult learners also face the dual challenge of developing basic literacy skills which includes decoding, comprehending, and producing print, along with English proficiency, represent different nationalities and cultural backgrounds \cite{wrigley2003language}. Hence, illiteracy is shown as a significant barrier that results in a person's struggling in every aspect of his or her daily life activity.

While there have not been any solutions to completely solve the illiteracy problem, recent developments of data science and artificial intelligence have brought a great opportunity to study how to support low-literate people in their lives. In this work, we propose SimplerVoice: a system that is able to generate key messages, and visual description for illiteracy. SimplerVoice could present easier-to-understand representations of complex objects to low-literate adult users, which helps them gain more confidence in navigating their own daily lives.

While the recent technology such as Google Goggles, Amazon's Flow, etc. proposed methods to parse the complex objects using image recognition, augmented reality techniques into the objects names, then to search for URLs of the objects information, the main challenges of SimplerVoice are to generate and retrieve simple, yet informative text, and visual description for illiterate people. This includes supporting adult basic education (ABE), and the English as a second language acquisition (SLA) training by performing natural language processing, and information retrieval techniques, such as: automatically generating sensible texts, word-sense-disambiguation and image-sense-disambiguation mechanism, and retrieving the optimal visual components. We propose the overall framework, and demonstrate the system in a case study of grocery shopping where SimplerVoice generates key text, and visual manual of how to use grocery products. The system prototype are also provided, and the empirical evaluation shows that SimplerVoice is able to provide users with simple text, and visual components which adequately convey the product's usage.

The organization of the paper is as follows. First, we have a quick review of previous works of text-to-image synthesis field in Section \ref{sec:related}. In Section \ref{sec:system}, we show our system design, including 4 parts as Input Retrieval, Object2Text, Text2Visual, and Output Display, along with the challenges of each components, and the proposed solution. We report the empirical evaluation of the proposed methods using real-world datasets for a case study in Section \ref{sec:evaluation}. Finally, Section \ref{sec:conclusion} concludes this paper, and states future work directions.
\begin{figure*}[htb!]
\includegraphics[width=\linewidth]{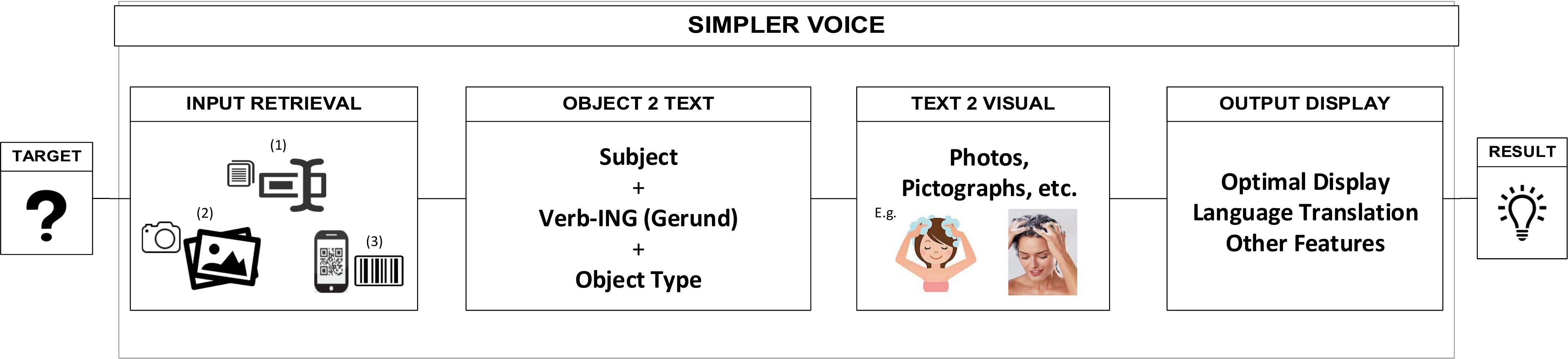}
\caption{Overall system structure \& workflow of SimplerVoice, with 4 main components: input retrieval, object2text, text2visual, and output display.}
\label{fig:system}
\end{figure*}

\section{Related Work}
\label{sec:related}
In the field of ABE and SLA, researchers have conducted a number of studies to assist low-literate learners in their efforts to acquire literacy and language skills by reading interventions, and providing specific instructions through local education agencies, community colleges and educational organizations \cite{condelli2004identifying,condelli2010impact}.

In augmentative and alternative communication (AAC) study, text-to-picture systems were proposed in \cite{widgit,goldberg2009toward}. \cite{widgit} used a lookup table to transliterate
each word in a sentence into an icon which resulted in a sequence of icons. Because the resulting icons sequence might be difficult to comprehend, the authors in \cite{goldberg2009toward} introduced a system using a concatenative or ''collage'' approach to select and display the pictures corresponding to the text.

To generate images from text, the authors  in \cite{schroff2011harvesting} proposed an approach to automatically generate a large number of images for specified object classes that downloads all contents from a Web search query, then, removes irrelevant components, and re-ranks the remainder. However, the study did not work on action-object interaction classes, which might be needed to describe an object.

Another direction is to link the text to a database of pictographs. \cite{vandeghinste2015translating} introduced a text-to-pictograph translation system that is used in an on-line platform for augmentative and alternative communication. The text-to-pictograph was built, and evaluated on email text messages. Furthermore, an extended study of this work was provided in \cite{sevens2016improving} which improved the Dutch text-to-pictograph through word sense disambiguation.

Recently, there have been studies that proposed to use deep generative adversarial networks to perform text-to-image synthesis \cite{reed2016generative,zhang2016stackgan}. However, these techniques might still have the limitation of scalability, or image resolution restriction.

\section{System Design}
\label{sec:system} 

In this section, we describe the system design, and workflow of SimplerVoice (Figure \ref{fig:system}). SimplerVoice has 4 main components: input retrieval, object2text, text2visual, and output display. Figure \ref{fig:system} provides the overall structure of SimplerVoice system.

\subsection{Overview}
\label{sec:system-overview}
Given an object as the target, SimplerVoice, first, retrieves the target input in either of 3 representations: (1) object's title as text, (2) object's shape as image, or (3) other forms, e.g.\ object's information from scanned barcode, speech from users, etc. Based on the captured input, the system, then, generates a query string/sequence of text which is the key message describing the object's usage. Due to low-literates' lack of reading capability, the generated text requires not only informativeness, but also simplicity, and clarity. Therefore, we propose to use the "S-V-O" query's canonical representation as below:
\begin{center}
[\textit{\textbf{S}ubject}] + [\textit{\textbf{V}erb-ing}] + \textit{(with)} + [\textit{\textbf{O}bject Type/Category}]
\end{center}
The intuition of this query representation is that the generated key message should be able to describe the action of a person using, or interacting with the target object. Moreover, the simple "S-V-O" model has been proposed to use in other studies \cite{yang2011corpus,krishnamoorthy2013generating,guadarrama2013youtube2text} since it is able to provide adequate semantics meaning. The detail of generating the S-V-O query is provided in Section \ref{sec:system-o2t}.

Once the query is constructed, SimplerVoice converts the query text into visual forms. There is a variety of visual formats to provide users: photos, icons, pictographs, etc. These visual components can be obtained by different means, such as: using search engine, mapping query/ontology to a database of images. However, the key point is to choose the optimal display for illiteracy which is described in Section \ref{sec:system-t2v}.  The result of SimplerVoice is provided further in Section \ref{sec:evaluation}.

\subsection{Object2Text}
\label{sec:system-o2t}
\begin{figure}
\includegraphics[width=1\linewidth]{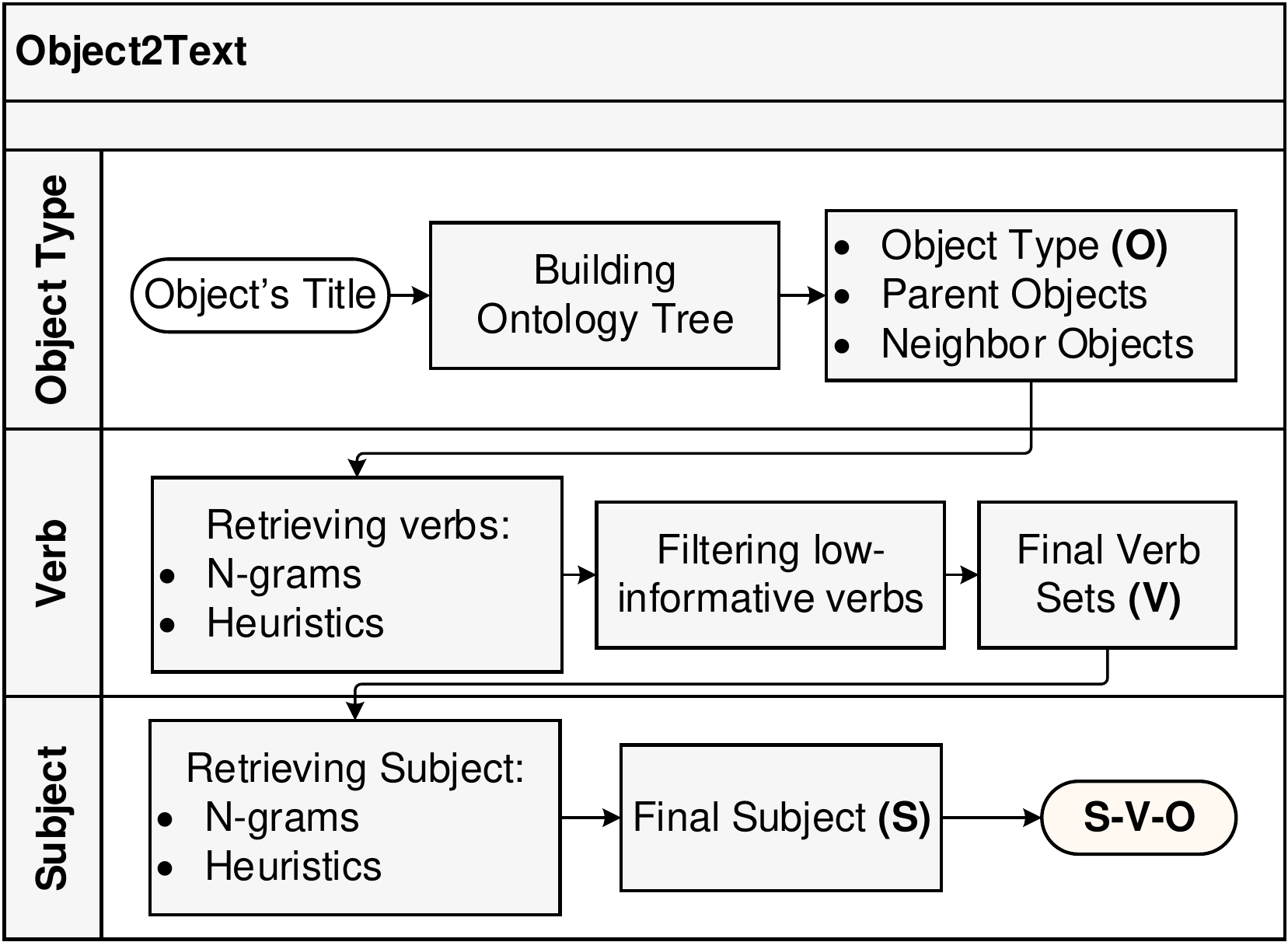}
\caption{The process of generating S-V-O in Object2Text.}
\label{fig:ob2txt}
\end{figure}
This section discusses the process of generating key message from the object's input. Based on the retrieved input, we can easily obtain the object's title through searching in database, or using search engine; hence, we assume that the input of object2text is the object's title.
The workflow of object2text is provided in Figure \ref{fig:ob2txt}. S-V-O query is constructed by the 3 steps below.
\subsubsection{Finding Object Type (O)}
\label{sec:system-o2t-o}
\begin{figure}
\includegraphics[width=1\linewidth]{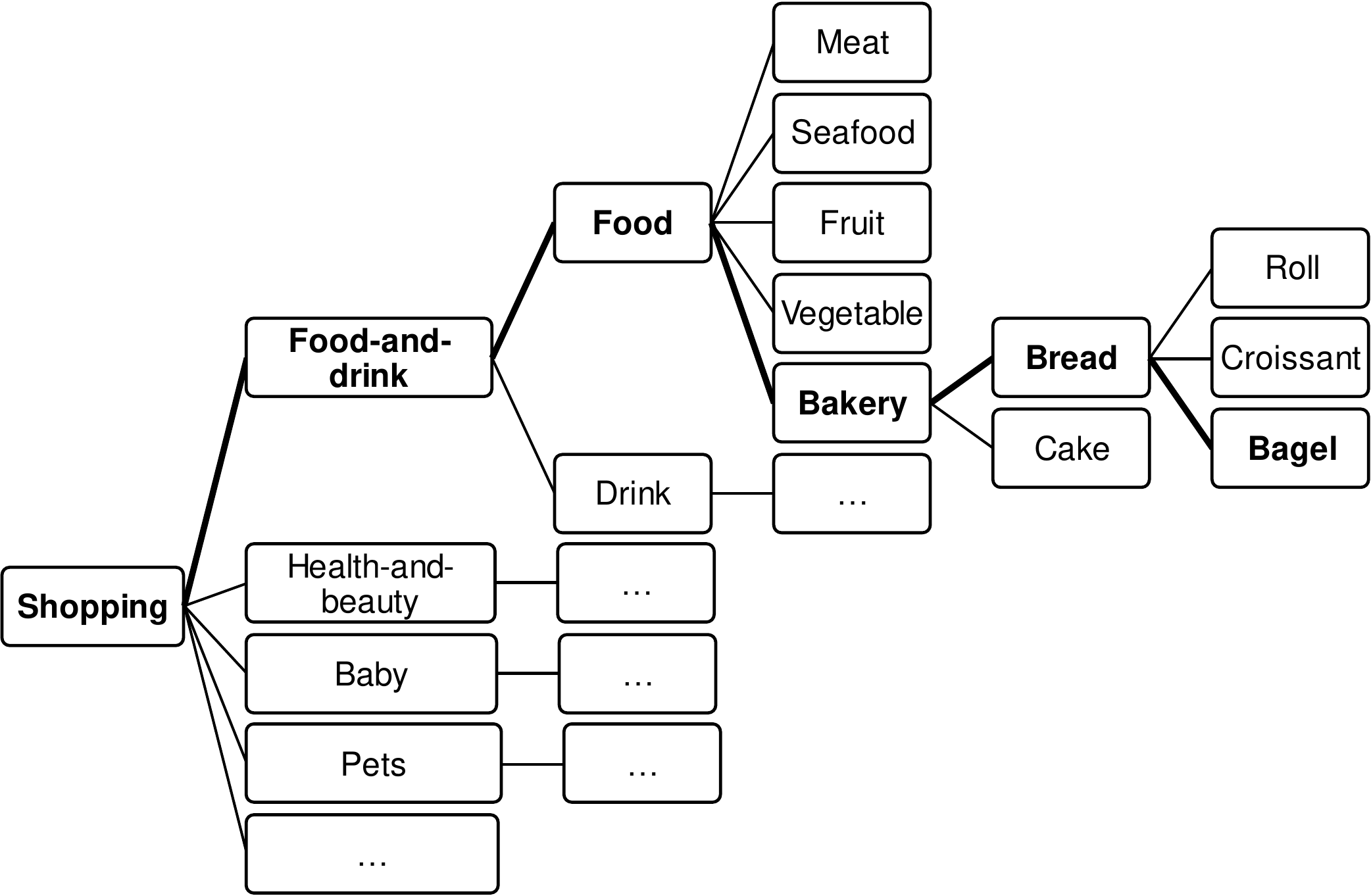}
\caption{An example of ontology sub-tree for "Bagel" built by crawled data from grocery shopping websites.}
\label{fig:ontology}
\end{figure}

In order to find the object type, SimplerVoice, first, builds an ontology-based knowledge tree. Then, the system maps the object with a tree's leaf node based on the object's title. For instance, given the object's title as ``\textit{Thomas' Plain Mini Bagels}", SimplerVoice automatically defines that the object category is ``\textit{bagel}". Note that both the knowledge tree, and the mapping between object and object category are obtained based on text-based searching / crawling web, or through semantic webs' content. Figure \ref{fig:ontology} shows an example of the sub-tree for object category "\textit{bagel}". While the mapped leaf node is the O in our S-V-O model, the parents nodes describe the more general object categories, and the neighbors indicate other objects' types which are similar to the input object. All the input object's type, the direct parents category, and the neighbors' are, then, put in the next step: generating verbs (V).

\subsubsection{Generating + Refining Verb Sets (V)}
\label{sec:system-o2t-v}
We propose to use 2 methods to generate the suitable verbs for the target object: heuristics-based, and n-grams model.
In detail, SimplerVoice has a set of rule-based heuristics for the objects. For instance, if the object belongs to a "\textit{food | drink}" category, the verb is generated as "\textit{eat | drink}". Another example is the retrieved "\textit{play}" verb if input object falls into "\textit{toy}" category. However, due to the complexity of object's type, heuristics-based approach might not cover all the contexts of object.
As to solve this, an n-grams model is applied to generate a set of verbs for the target object. An n-gram is a contiguous sequence of n items from a given speech, or text string. N-grams model has been extensively used for various tasks in text mining, and natural language processing field \cite{lin2003automatic,wang2007topical}. Here, we use the Google Books n-grams database \cite{michel2011quantitative,lin2012syntactic} to generate a set of verbs corresponding to the input object's usage. Given a noun, n-grams model can provide a set of words that have the highest frequency of appearance followed by the noun in the database of Google Books. For an example, "eaten", "toasted", "are", etc. are the words which are usually used with "bagel". To get the right verb form, after retrieving the words from n-grams model, SimplerVoice performs word stemming \cite{lovins1968development} on the n-grams' output.\\

\textit{Word-sense disambiguation}: In the real-world case, a word could have multiple meanings. This fact may affects the process of retrieving the right verb set. Indeed, word-sense disambiguation has been a challenging problem in the field of nature language processing. An example of the ambiguity is the object "cookie". The word "cookie" has 2 meanings: one is "a small, flat, sweet food made from flour and sugar" (context of biscuit), another is "a piece of information stored on your computer about Internet documents that you have looked at" (context of computing). Each meaning results in different verb lists, such as: "eat", "bake" for biscuit cookie, and "use", "store" for computing cookie. In order to solve the ambiguity, we propose to take advantage of the built ontology tree.

In detail, SimplerVoice uses the joint verb set of 3 types of nouns: the input object, the parents, and the neighbors as the 3 noun types are always in the same context of ontology. Equation \ref{eq:disambiguation} shows the word-sense disambiguation mechanism with $V$(Object) indicates the verb set of an object generated by heuristics, and n-grams model:
\begin{equation}
\label{eq:disambiguation}
\begin{split}
V_{final}(Object) = V(Object) \bigcap V(Parents) \\ \bigcap V(Neighbors)
\end{split}
\end{equation}
 
\begin{figure}
\includegraphics[width=1\linewidth]{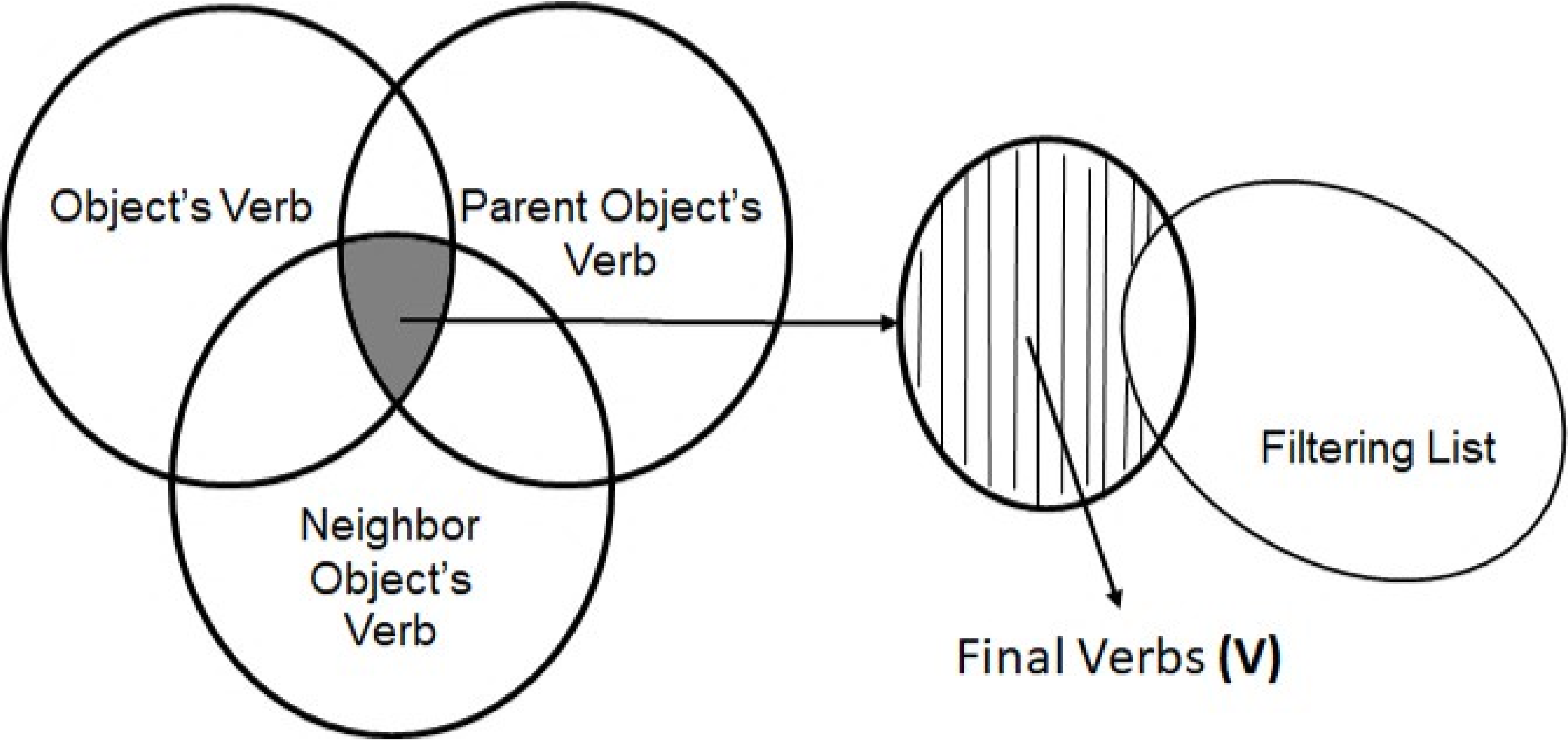}
\caption{Word-sense disambiguation, and low-informative verbs filtering.}
\label{fig:filtering}
\end{figure}

\textit{Low-informative verbs}: In order to ensure the quality of generated verbs, SimplerVoice maintains a list of restricted verbs that need to be filtered out. There are a lot of general, and low-informative verbs generated by n-grams model such as "be", "have", "use", etc. as these verbs are highly used in daily sentences/conversation. The restricted verb list could help to ensure the right specificity aspect. Hence, we modify (\ref{eq:disambiguation}) into (\ref{eq:filtering}). The process of word-sense disambiguation, and low-informative verb filtering is provided in Figure \ref{fig:filtering}:

\begin{equation}
\label{eq:filtering}
\begin{split}
V_{final}(Object) = (V(Object) \bigcap V(Parents) \\ \bigcap V(Neighbors))
\setminus Fil.List.
\end{split}
\end{equation}

\subsubsection{Defining Subject (S)}
\label{sec:system-o2t-s}
The approach to generate the subject (S) is similar to the verb (V). SimplerVoice also uses heuristics, and n-grams model to find the suitable actor S. In regard to heuristics method, we apply a rule-based method via the object's title, and object's category to generate S since there are objects only used by a specific group of S. For an example, if the object's title contains the word "woman, women", the S will be "Woman"; of if the object belongs to the "baby" product category, the S will be "Baby". Additionally, n-grams model also generates pronouns that frequently appear with the noun O. The pronouns output could help identify the right subject S, e.g. "she" - "woman, girl", "he" - "man, boy", etc. If there exists both "she", and "he" in the generated pronoun set, the system picks either of them.

\subsection{Text2Visual}
\label{sec:system-t2v}
Once the S-V-O is generated, Text2Visual provides users with visual components that convey the S-V-O text meanings.
\subsubsection{Image Sense Ambiguity} One simple solution to perform Text2Visual is to utilize existing conventional Web search engines. SimplerVoice retrieves top image results using S-V-O as the search query. However, there could be image sense ambiguity in displaying the result from search engine. For instance, if the object is "Swiss Cheese", user might not distinguish between "Swiss Cheese", and the general "Cheese" images. To solve the image sense ambiguity issue, the authors in \cite{goldberg2009toward} suggests to display multiple images to guide human perception onto the right target object's meaning.

\subsubsection{Optimal Visual Component} 
\begin{figure}
\centering
\includegraphics[width=0.5\linewidth]{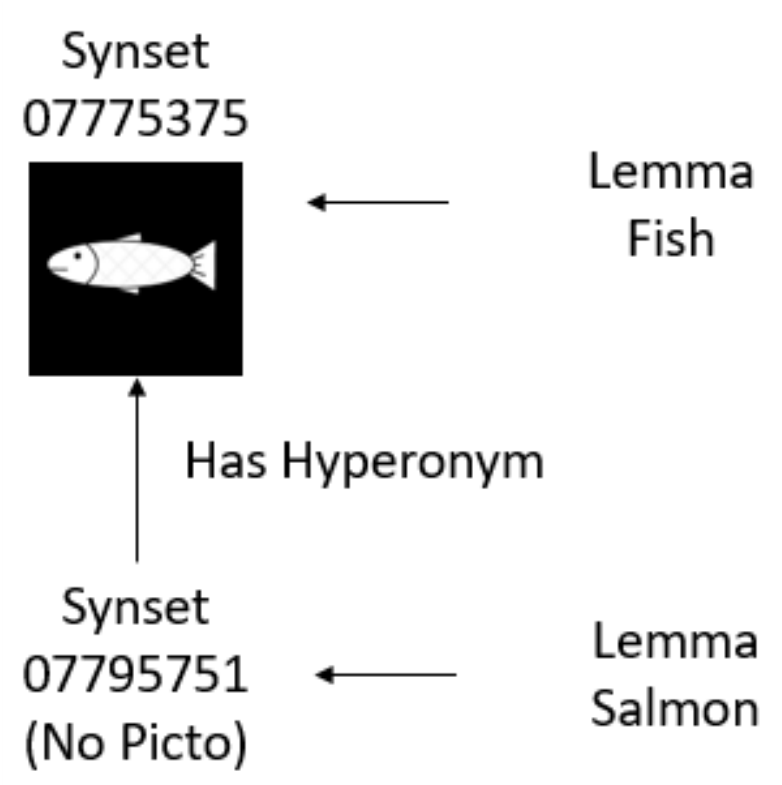}
\caption{Linking pictographs to words in S-V-O\protect\cite{vandeghinste2015translating}}
\label{fig:picto}
\end{figure}

Additionally, since SimplerVoice is designed for illiteracy, the system needs to display the optimal visual component suitable for low-literate people. In \cite{medhi2007optimal}, the authors study the effectiveness of different types of audio-visual representations for illiterate computer users. While there is no difference between dynamic and static imagery (mixed result in different use cases), hand-drawn or cartoons are shown to be easier for low-literate users to understand than photorealistic representations. Therefore, SimplerVoice also provides users with pictographs display along with images. We use the Sclera database of pictographs \cite{scl2017}. Each S-V-O word is mapped with a corresponding Sclera pictograph file. The detail of how to perform the mapping is discussed in \cite{vandeghinste2015translating}. Intuitively, the process is described as: first, the system manually links a subset of words with pictographs' filenames; then, if the manual link is missing, the word is linked to the close synset using WordNet (Figure \ref{fig:picto}).

\section{Evaluation}
\label{sec:evaluation}
In this section, we demonstrate the effectiveness of SimplerVoice system in a case study of grocery shopping. The section organization is as follows: first, we describe the real dataset, and setup that SimplerVoice uses; second, we provide the prototype system which is a built application for end-users; finally, we show the results of SimplerVoice along with users feedback.

\subsection{Case Study}
\label{sec:evaluation-dataset}
In the case study of grocery products shopping, we use a database of $114,522$ products' description crawled from multiple sources. Each product description contains 4 fields: UPC code, product's title, ontology path of product category, and URL link of the product. Since it is recommended to utilize various devices of technology, such as computers or smart phones in adult ESL literacy education \cite{choi2015literacy}, we build a mobile application of SimplerVoice for illiterate users. The goal of SimplerVoice is to support users with key message, \& simple visual components of how to use the products given the scanned barcode (UPC code), or products' name retrieved from parsing products images taken by end-users' phone cameras. Section \ref{sec:evaluation-prototype} shows our SimplerVoice application description.

\subsection{Prototype System}
\label{sec:evaluation-prototype}
\begin{figure}
\centering
\includegraphics[width=1\linewidth]{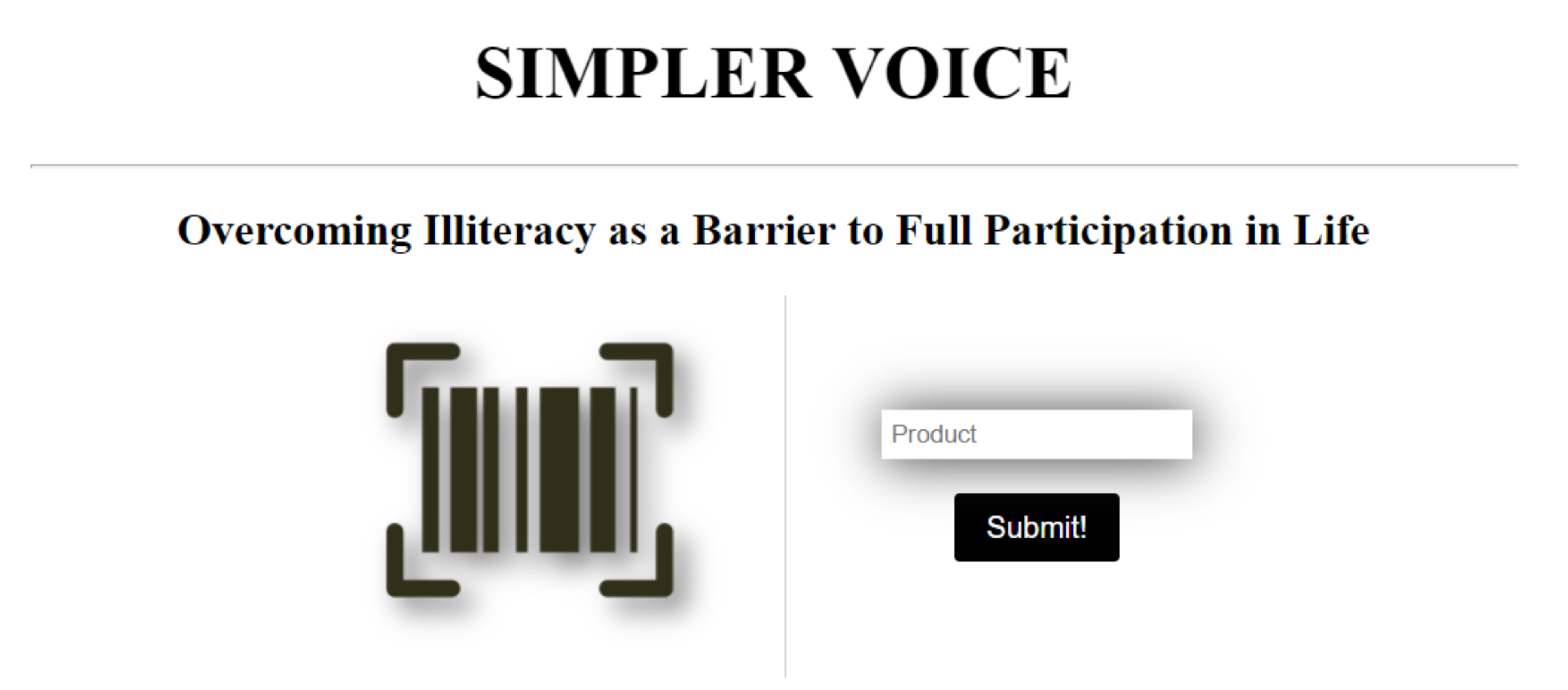}
\caption{Retrieving object's input of grocery product through text, or scanned image.}
\label{fig:demo-input}
\end{figure}

There are 2  means to retrieve the object's input through SimplerVoice application: text filling, or taking photos of barcode / products' labels (Figure \ref{fig:demo-input}). SimplerVoice automatically reads the target grocery product's name, and proceeds to the next stage.

\begin{figure}
\centering
\includegraphics[width=1\linewidth]{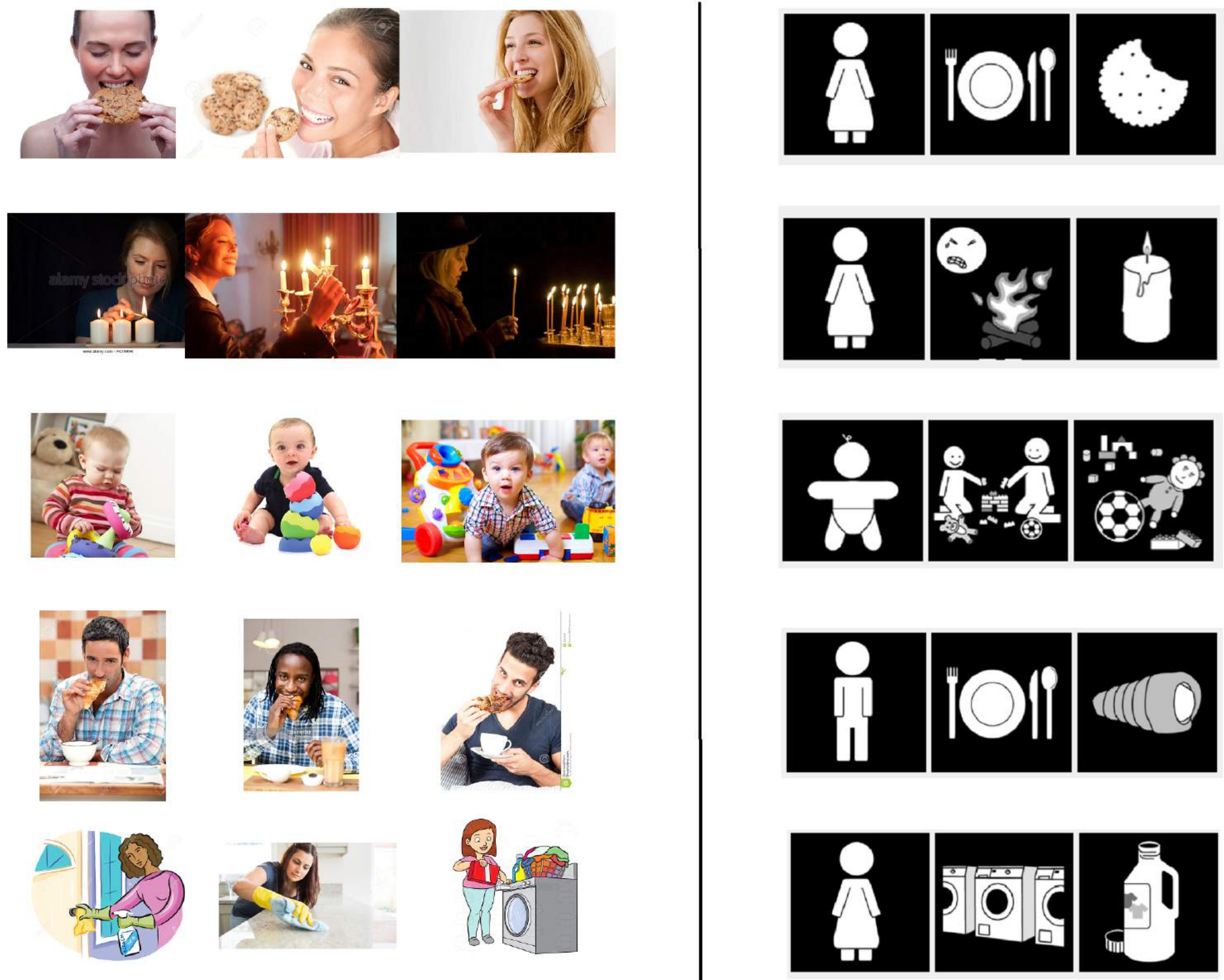}
\caption{Examples of visual components generated by SimplerVoice corresponding to products in Table \ref{tab:object2text}. Left column is the photorealistic image, and right column is the generated pictograph.}
\label{fig:demo-example}
\end{figure}

\begin{figure*}
\centering
\includegraphics[width=0.9\linewidth]{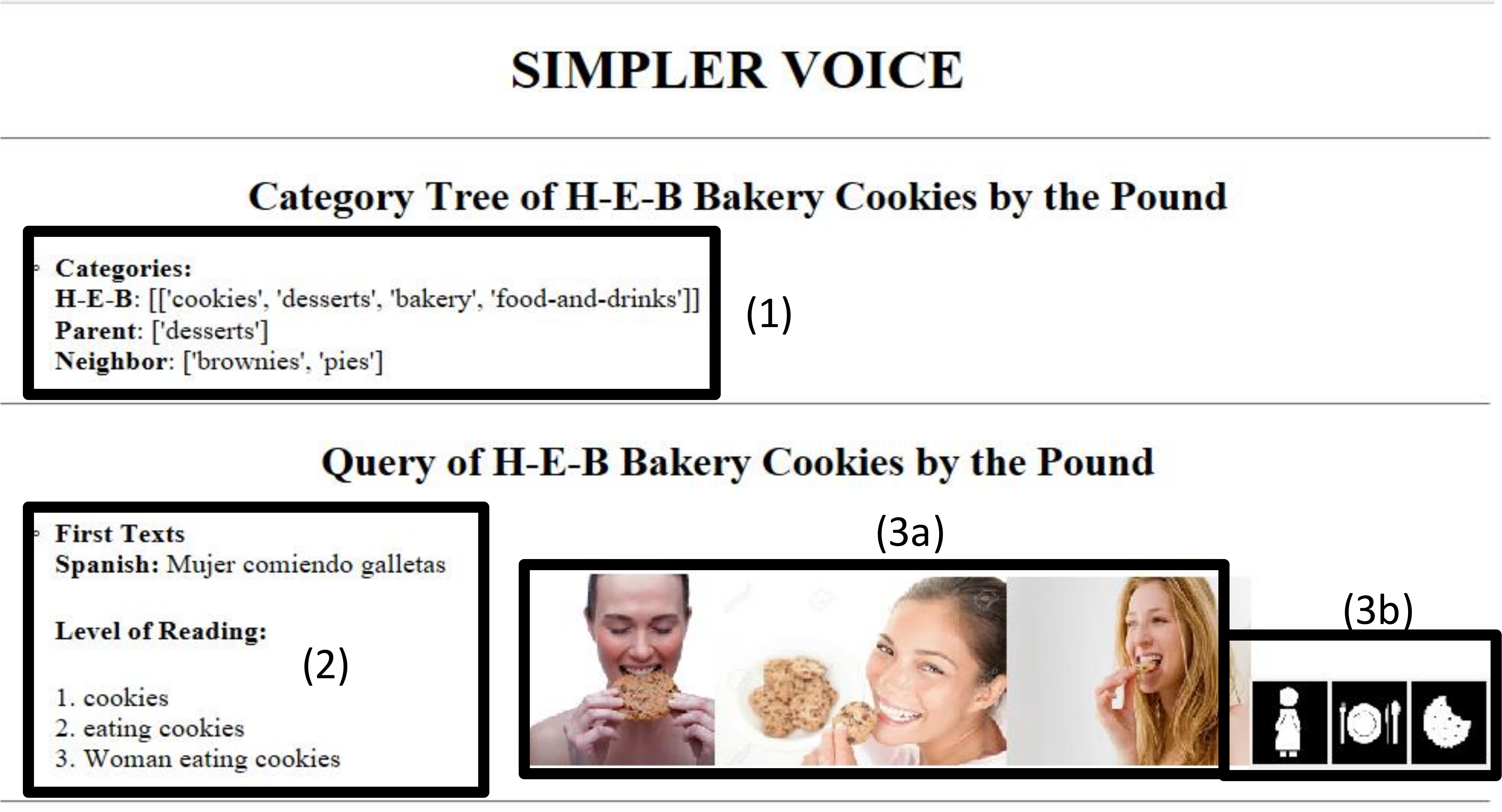}
\caption{An example of key messages \& visual components generated by SimplerVoice: (1) The product description of "H-E-B Bakery Cookies by the Pound", (2) Generated key messages, (3) Visual components of the products: photorealistic images (3a) + pictographs (3b).}
\label{fig:demo-result}
\end{figure*}

Based on the built-in ontology tree, SimplerVoice, then, finds the object's category, the parent, and the neighboring nodes. The next step is to generate the S-V-O message (e.g. Table \ref{tab:object2text}), and visual description (e.g. Figure \ref{fig:demo-example}) of product's usage. Figure \ref{fig:demo-result} shows an example of the result of SimplerVoice system for product "H-E-B Bakery Cookies by the Pound" from a grocery store: (1) the product description, (2) key messages, and (3) visual components. The product description includes the product's categories searched on the grocery store's website \cite{heb}, the parent node's, and the neighbors - similar products' categories. The S-V-O query, or key message for "H-E-B Bakery Cookies by the Pound" is generated as "Woman eating cookies". Additionally, we support users with language translation into Spanish for convenience, and provides different levels of reading. Each reading level has a different level of difficulty: The higher the reading level is, the more advanced the texts are. The reason of breaking the texts into levels is to encourage low-literate users learning how to read. Next to the key messages are the images, and pictographs.

\begin{table}[tbp]
\centering
\caption{Examples of generated key messages from objects' titles}
\label{tab:object2text}
\begin{tabular}{|l|l|}
\hline
\multicolumn{1}{|c|}{\textbf{Object Title}}                                                        & \multicolumn{1}{c|}{\textbf{Key Message}}                           \\ \hline
\begin{tabular}[c]{@{}l@{}}H-E-B Bakery Cookies by \\ the Pound\end{tabular}                       & Woman eating cookie                                                 \\ \hline
Culpitt                                                                                            &  Woman lighting candle                                                    \\ \hline
\begin{tabular}[c]{@{}l@{}}Fisher-Price Brilliant Basics\\ Rock-a-Stack (6-36 Months)\end{tabular} & Baby playing toy                                                \\ \hline
Cafe Valley Cocktail Croissant                                                                     & Man eating croissant                                                \\ \hline
Clorox                                                                                             & \begin{tabular}[c]{@{}l@{}}Woman washing with\\ bleach\end{tabular} \\ \hline
\end{tabular}
\end{table}

\begin{table*}[]
\centering
\caption{The MOS scores of all approaches}
\label{tab:mos}
\begin{tabular}{|c|l|c|c|c|}
\hline
\textbf{Approach}     & \multicolumn{1}{c|}{\textbf{\begin{tabular}[c]{@{}c@{}}Approach Description\end{tabular}}}                                       & \textbf{\begin{tabular}[c]{@{}c@{}}MOS Score Range\end{tabular}} & \textbf{\begin{tabular}[c]{@{}c@{}}Mean MOS Score\end{tabular}} & \textbf{Stdev} \\ \hline
\textbf{Baseline 1}   & \begin{tabular}[c]{@{}l@{}}1. Product title\\ 2. Original product package\end{tabular}                                                    & 1 - 4.25                                                            & 2.57                                                               & 1.17           \\ \hline
\textbf{Baseline 2}   & \begin{tabular}[c]{@{}l@{}}1. Product title\\ 2. Top product images using search engines:\\ Google, and Bing\end{tabular}                    & 1 - 4.7                                                             & 2.86                                                               & 1.27           \\ \hline
\textbf{SimplerVoice} & \begin{tabular}[c]{@{}l@{}}1. SimplerVoice generated text \\ 2. SimplerVoice generated visual components:\\ photorealistic images, pictographs\end{tabular} & 3.75 - 5                                                            & 4.82                                                               & 0.35           \\ \hline
\end{tabular}
\end{table*}

\begin{figure*}
\centering
\includegraphics[width=0.7\linewidth]{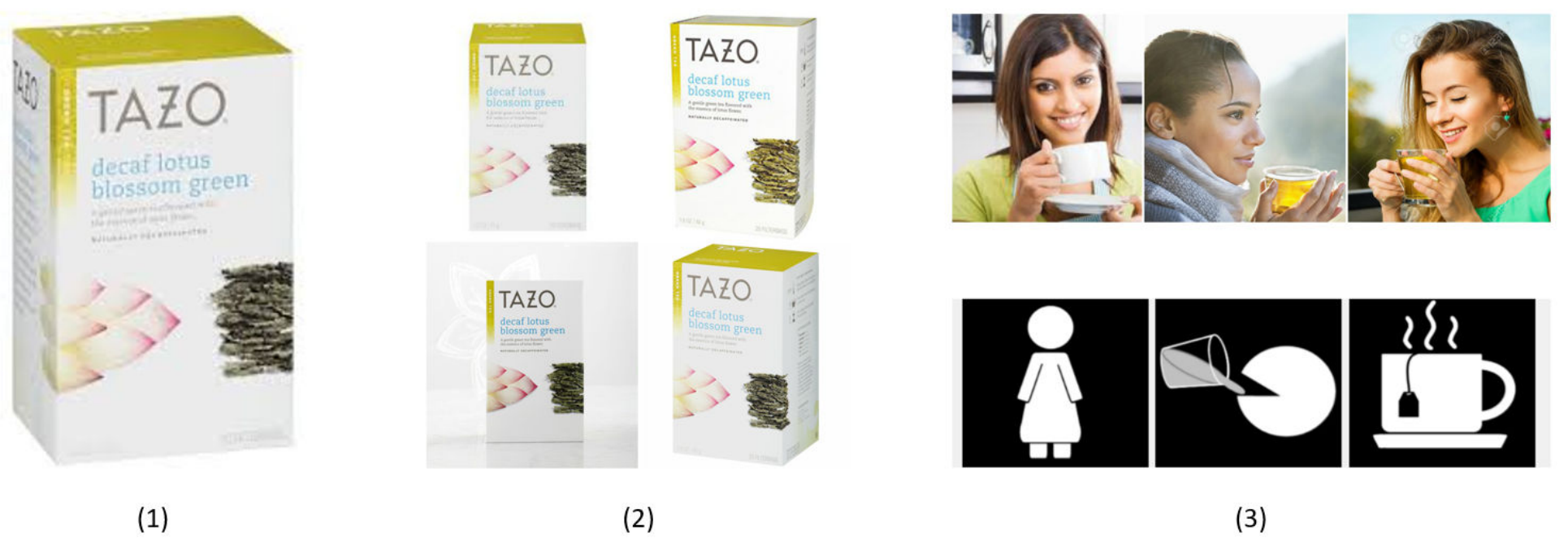}
\caption{3 approaches to generate the product description for product "Tazo Decaf Lotus Blossom Green Tea Filterbags": (1) Baseline 1 - Using the original package and product title, (2) Baseline 2 - Using the result from search engine as product description, and (3) SimplerVoice visual description.}
\label{fig:baselines-method}
\end{figure*}

\subsection{Experiment}
\subsubsection{Experimental Setup \& Dataset}
To evaluate our system, we compared SimplerVoice to the original product description / package (baseline 1) and the top images result from search engines of the same product (baseline 2). Given a set of products, we generated the key message \& visual description of each product using 3 approaches below. An example of the 3 approaches is provided in Fig. \ref{fig:baselines-method}.

\begin{itemize}
\item \textit{Baseline 1}: We captured and displayed the product package photos and the product title text as product description.
\item \textit{Baseline 2}: The product description was retrieved by search engine using the product titles, and then presented to the users as the top images result from Google and Bing. We also provided the product title along with the images.
\item \textit{SimplerVoice}: We shown the generated key messages (Tab. \ref{tab:object2text}), and visual description including 2 components: photorealistics images and pictographs (Fig.\ref{fig:demo-example}) from SimplerVoice system.
\end{itemize}

\begin{figure}
\centering
\includegraphics[width=0.8\linewidth]{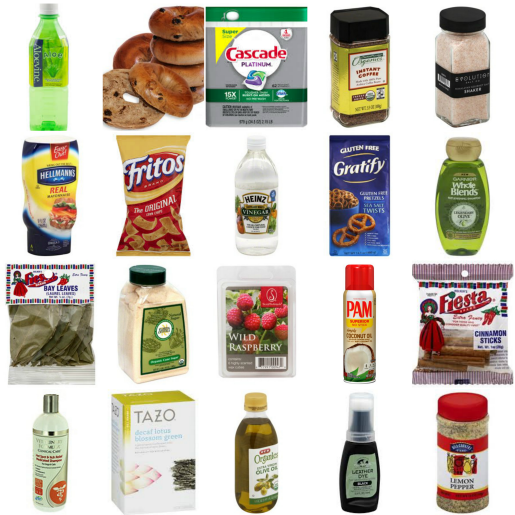}
\caption{A sample of 20 random products  from the U.S. grocery store used in our experiment \protect\cite{heb}.}
\label{fig:sample}
\end{figure}

Intuitively, baseline 1 shows how much information a user would receive from the products' packages without prior knowledge of the products while baseline 2 might provide additional information by showing top images from search engines. With the baseline 2, we attempt to measure whether merely adding "relevant" or "similar" products' images would be sufficient to improve the end-users' ability to comprehend the product's intended use. Moreover, with SimplerVoice, we test if our system could provide users with the proper visual components to help them understand the products' usage based on the proposed techniques, and measure the usefulness of SimplerVoice's generated description.

We evaluated the effectiveness \& interpretability of 3 above approaches by conducting a controlled user study with 15 subjects who were Vietnamese native and did not speak/comprehend English. A dataset of random 20 U.S. products including products' title, UPC code, and product package images were chosen to be displayed in the user study. Note that the 15 participated subjects had not used the 20 products before and were also not familiar with the packaged products including the chosen 20 products; hence, they were "illiterate" in terms of comprehending English and in terms of having used any of the products although they might be literate in Vietnamese.

Each participated user was shown the product description generated from each approach, and was asked to identify what the products were and how to use them. The users' responses were then recorded in Vietnamese and were assigned to a score if they "matched" the correct answer by 3 experts who were bilingual in English and Vietnamese. In this study, we used the "mean opinion score" (MOS) \cite{recommendation2006vocabulary,streijl2016mean} to measure the effectiveness: how similar a response were comparing to the correct product's usage. The MOS score range is 1-5 (1-Bad, 2-Poor, 3-Fair, 4-Good, 5-Excellent) with 1 means incorrect product usage interpretability - the lowest level of effectiveness and 5 means correct product usage interpretability - the highest effectiveness level. The assigned scores corresponding to responses were aggregated over all participated subjects and over the 3 experts. The result of the score is reported in the next section Result.

\subsubsection{Result}
\label{sec:result}
Table \ref{tab:mos} shows the MOS scores indicating the performance of 3 approaches. The mean of MOS scores of baseline 1 is the lowest one: 2.57 (the standard deviation (stdev) is 1.17), the baseline 2 mean score is slightly higher than the baseline 1's: 2.86 (the stdev is 1.27), while SimplerVoice evaluation score is the highest one: 4.82 (the stdev is 0.35) which means the most effective approach to provide users with products' usage. Additionally, a paired-samples t-test was conducted
to compare the MOS scores of users' responses among all
products using baseline 1 and SimplerVoice system. There was a significant difference in the scores for baseline 1 (Mean = 2.57, Stdev = 1.17) and SimplerVoice (Mean = 4.82, Stdev = 0.35); t= -8.18224, p =1.19747e-07. These results show that there is a statistically significant difference in the MOS means between baseline 1 and SimplerVoice and that SimplerVoice performs more effectively than baseline 1 over different types of products.

Baseline 1 scores ranges from 1 to 4.25 over all products as some products are easily to guess based on product package images, such as bagels, pretzels, soda, etc. while some products packages might cause confusion, such as shoe dye, wax cube, vinegar, etc. For an example, all participated users were able to recognize the "Always Bagels Cinnamon Raisin Bagels" product as "a type of bread" and its usage as "eating" using baseline 1 while the "ScentSationals Wild Raspberry Fragrance Wax Cubes" product were mostly incorrectly recognized as a type of "candy" for "eating".

Baseline 2 scores range over all products is 1 - 4.7. The baseline 2 has higher score than baseline 1 since users were provided more information with the top result product images from search engine. For instance, given the "Fiesta Cinnamon Sticks" product, most users' responses were "a type of pastries - cannoli" for "eating" based on baseline 1. Since baseline 2 provided more photos of cinnamon sticks without the packaging, the users were able to recognize the products as cinnamon. Moreover, the score of baseline 2 is only slightly higher than baseline 1 because search engines mostly return similar images from product package, hence, might provide only  little additional information to the participants.

SimplerVoice scores ranges from 3.75 to 5 which is higher than baseline 1, and baseline 2. SimplerVoice score has low standard deviation indicating the consistent effectiveness along different types of products. While performing the user study, we also notice that the culture differences is an important factor to the result. For an example, the product has lowest score is the "Heinz Distilled White Vinegar" since there were participated users who have never used vinegar before. These participants are from the rural Northern Vietnam area where people might have not known the vinegar product.

\section{Conclusion}
\label{sec:conclusion}
In this work, we introduce SimplerVoice: a key message \& visual description generator system for illiteracy. To our best knowledge, SimplerVoice is the first system framework to combine multiple AI techniques, particularly in the field of natural language processing, and information retrieval, to support low-literate users including low-literate ESL learners building confidence on their own lives, and to encourage them to improve their reading skills. Although awareness by itself does not solve the problem of illiteracy, the system can be put in different contexts for education goals. SimplerVoice might be a valuable tool for both educational systems, and daily usage.

The SimplerVoice system was evaluated and shown to achieve higher performance score comparing to other approaches. Moreover, we also introduced the SimplerVoice mobile application and have the application used by participants in the Literacy Coalition of Central Texas's SPARK program \cite{spark2017}. We received positive end-users' feedback for the prototype, and plan to add more features for SimplerVoice.

One of the future work is to improve the input retrieval of the system, so that SimplerVoice can automatically recognize the object through the object's shape. Another direction is to extend the work in other different real-world use cases, and demonstrate its effectiveness on those case studies.

\section*{Acknowledgments}
This research was conducted under the auspices of the IBM Science for Social Good initiative. The authors would like to thank Christian O. Harris and Heng Luo for discussions.

\balance{}

\bibliographystyle{SIGCHI-Reference-Format}
\bibliography{C:/Users/MinhNguyen/Downloads/bloomberg/ref}

\end{document}